\documentclass[10pt,twocolumn,letterpaper]{article}
\pdfoutput=1
\usepackage{times}
\usepackage{epsfig}
\usepackage{graphicx}
\usepackage{amsmath}
\usepackage{amssymb}
\usepackage{bm,pifont}

\usepackage[pagebackref=true,breaklinks=true,letterpaper=true,colorlinks,bookmarks=false]{hyperref}

\begin{document}

\title{Multi-Semantic Interactive Learning for Object Detection}

\author{Shuxin Wang${^{1,2}}$  \quad \quad
Zhichao Zheng${^{2,3}}$\quad \quad
Yanhui Gu${^{1,2}}$  \quad \quad
Junsheng Zhou${^{1,2}}$  \quad  \quad
Yi Chen${^{1,2}}$\thanks{Corresponding author.}\\
 $^2$School of Computer and Electronic Information, Nanjing Normal University, China\\ 
$^1$\{\tt\small{202243036}, gu,  zhoujs, cs\underline{ }chenyi\}@njnu.edu.cn  \quad
 $^3$zheng\underline{ }zhichaoX@163.com  \\
}

\maketitle

\begin{abstract}
Single-branch object detection methods use shared features for localization and classification, yet the shared features are not fit for the two different tasks simultaneously. Multi-branch object detection methods usually use different features for localization and classification separately, ignoring the relevance between different tasks. Therefore, we propose multi-semantic interactive learning (MSIL) to mine the semantic relevance between different branches and extract multi-semantic enhanced features of objects. MSIL first performs semantic alignment of regression and classification branches, then merges the features of different branches by semantic fusion, finally extracts relevant information by semantic separation and passes it back to the regression and classification branches respectively. More importantly, MSIL can be integrated into existing object detection nets as a plug-and-play component. Experiments on the MS COCO, and Pascal VOC datasets show that the integration of MSIL with existing algorithms can utilize the relevant information between semantics of different tasks and achieve better performance.
\end{abstract}

\section{Introduction}

\begin{figure}[t]
\begin{center}
   \includegraphics[width=0.9\linewidth]{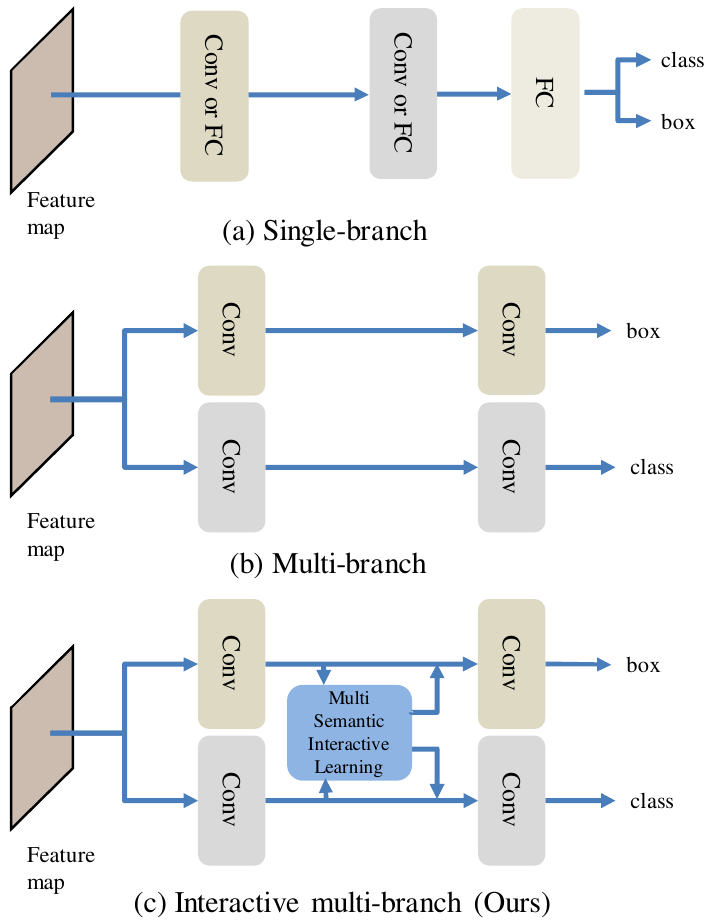}
\end{center}
   \caption{Architecture of detection head. (a) Single-branch detection head. Shared features are not fit for the two different tasks of regression and classification simultaneously. (b) Multi-branch detection head with independent branches, ignoring the relevance  between branches. (c) Ours, emphasizing both independence and relevance of branches.}
   \label{fig:introduction}
\end{figure}
As one of the most fundamental and challenging problems in computer vision, object detection has been widely applied in the areas of automatic driving~\cite{3}, text detection~\cite{4}, face detection~\cite{5}, etc. Object detection provides one of the most basic information to computer vision applications: What objects are where~\cite{1}? Therefore, object detection can be divided into two different tasks: localization and classification. Generally, localization can be viewed as the regression problem and classification can be viewed as the classification problem. 

In recent years, the rapid development of deep learning~\cite{2}has injected new vitality into object detection. Compared to earlier traditional methods, deep learning-based approaches have made significant breakthroughs.  Based on the number of branches, object detection methods can be divided into single-branch and multi-branch nets.

Single-branch nets are employed by many two-stage detectors~\cite{6,7,8,9} where shared detection head was used to predict position and class, which means that both regression and classification are computed in one branch (see Figure \ref{fig:introduction}-(a)). There are two main implementations, convolution of the extracted features, such as Faster Region Convolutional Neural Networks (R-CNN)~\cite{8}, and using fully connected layers for the features, such as Feature Pyramid Networks (FPN)~\cite{9}.

Using shared parameters for regression and classification introduces the problem of feature coupling~\cite{10}: features that are suitable for classification are not necessarily fit for regression. Similarly, features that are suitable for regression are not necessarily the most class discriminative features for that object.

Multi-branch nets were used by many one-stage detectors~\cite{11,12,13,14,15,16}, which use two branches to predict position and class separately (see Figure \ref{fig:introduction}-(b)) instead of using shared detection head. Separating localization regression and class classification makes each of them more focused on its own task.

The regression and classification branches are independent of each other, solving the problem of feature coupling, but ignoring the relevance between them. Currently, some methods attempt to re-establish the connection between regression and classification. TOOD~\cite{feng2021tood} aligns features through the detection boxes of regression and classification to reduce bias. But detection boxes are only a part of correlation, and their utilization of correlation is not sufficient. GFLv2~\cite{li2021generalized} and BorderDet~\cite{qiu2020borderdet} use regression information to enhance features. However, classification information is also important and should be utilized.

To utilize the relevance between branches in multi-branch nets, we proposed multi-semantic interactive learning for object detection (MSIL). MSIL interacts at the feature level, eliminating the limitations of manually designing interaction content. Additionally, MSIL utilizes both regression and classification information, enabling more comprehensive utilization of available information. 

MSIL first performs semantic alignment of features from different branches to extract common features. Then the semantic information in different branches are fused together in semantic fusion. Finally, semantic separation is performed, and the fused features are reduced to the original semantics to obtain relevant information. By utilizing the relevant information, the regression branch focuses more on the boundary of the object and the classification branch focuses more on the region where the object is located. The contributes of this paper are summarized as follows:
\begin{itemize}
\item We propose multi-semantic interactive learning (MSIL) for object detection. The relevant information between branches was utilized through semantic alignment, semantic fusion, and semantic separation. MSIL makes the regression branch focus more on object boundaries, and makes the classification branch focus on the region where the object is located. 
\item Multi-semantic interactive learning can be easily integrated into multi-branch nets, which has been integrated and achieved better performance on FCOS and VFNet~\cite{16}, respectively.
\item We show the impact of MSIL on regression and classification by using relevant information through visualization. And experiments on two datasets verify that MSIL can leverage the relevant information between semantics to improves the performance of different methods.
\end{itemize}

\begin{figure*}
\begin{center}
   \includegraphics[width=1\linewidth]{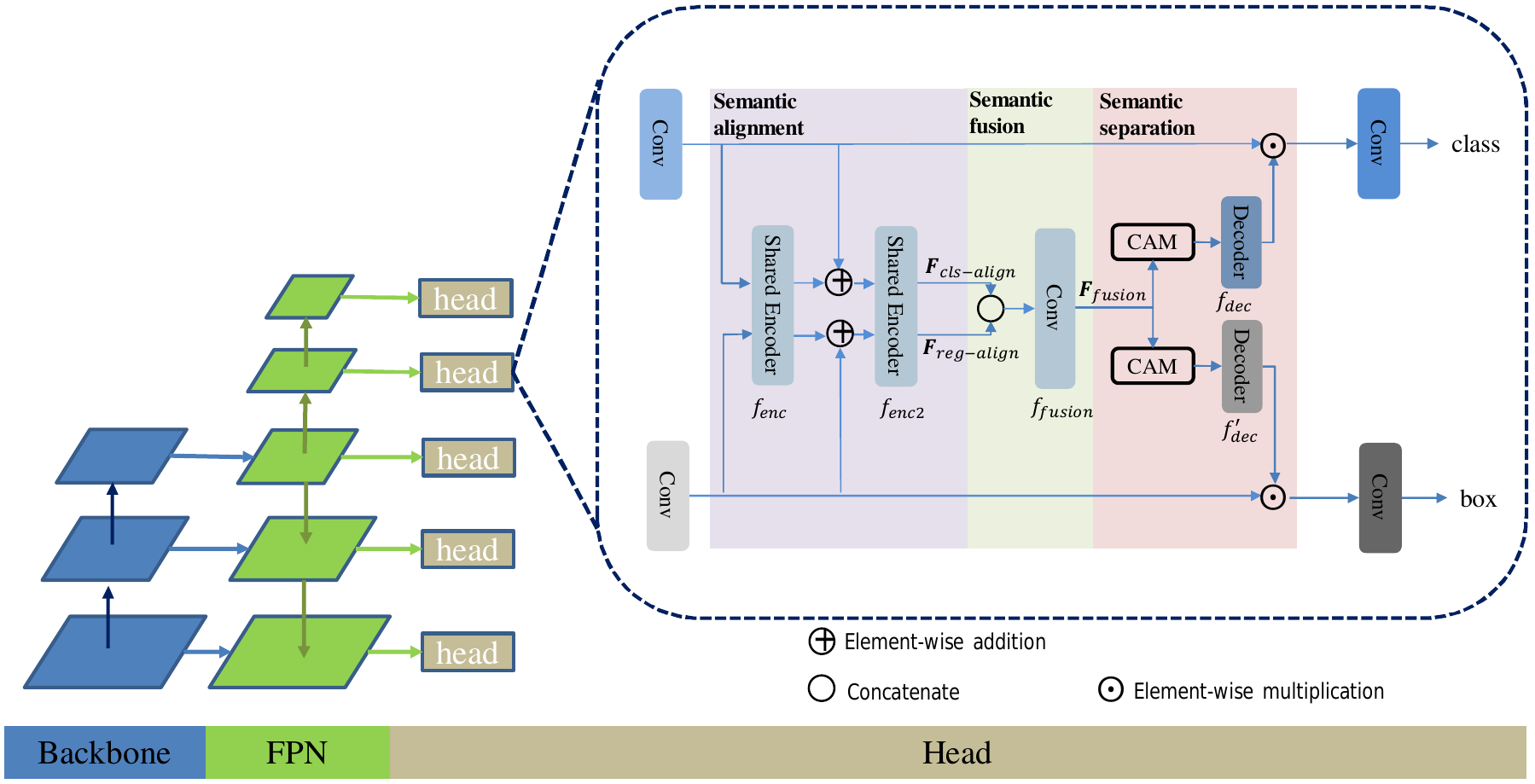}
\end{center}
   \caption{MSIL Network structure. The features of the classification and regression branches passes through the shared encoder $f_{enc}$ and adds to itself to obtain the aligned features $F_{cls-align}$ and $F_{reg-align}$. Then semantic fusion is performed to obtain $F_{fusion}$. $F_{fusion}$ was semantically separated and returned to the original semantics by the different channel attention module (CAM) and $f_{dec}$.}
   \label{fig:net}
\end{figure*}

\section{Related Works}
R-CNN~\cite{17} is the first deep learning-based object detection algorithm, which uses sliding windows to generate candidate boxes, and then extracts features in the candidate boxes for regression and classification. R-CNN significantly improves the performance of object detection by abandoning the traditional using manually designed features. R-CNN uses one branch to predict location and class, and it convolves the extracted features and feeds them to the fully connected layer before passing them to the regressor and classifier.

There are numerous improvements for the R-CNN series: SPP-Net~\cite{18} significantly improves the computational speed by reusing the extracted features. Fast R-CNN~\cite{6} further improves the efficiency by simplifying the pooling operation with RoI pool~\cite{6}, and integrating the regressor into the neural network. Faster R-CNN~\cite{8} is the paradigm of the R-CNN series, which proposed the Region Proposal Network (RPN) based on subsequent improvements. Discarding the previous selective search~\cite{19} algorithm, RPN uses neural networks to generate candidate boxes. It was followed by a series of improvements, such as Cascade R-CNN~\cite{20}, etc. It is worth mentioning that the R-CNN series has always used one branch to predict position and class. Despite various adjustments made by subsequent improvements, regression and classification have always been coupled together. Later, Double-Head~\cite{10} discussed in this regard.

YoloV1~\cite{15} is the first single-stage detector, which divides the image into individual lattices, and then predicts all lattices at once. It is worth mentioning that YoloV1 only uses points near the center of the object to predict edges. Because these points are thought to produce higher quality detections. However, this strategy leads to low recall of YoloV1. CornerNet~\cite{51} is another single-stage detector that detects a pair of corners of a bounding box and groups them to form the final detected bounding box.

Recently, FCOS~\cite{13} has attracted much attention. FCOS introducing "center-ness"~\cite{13}  to suppress low-quality boxes and solves the problem of low recall by using all points for prediction. Furthermore, FCOS uses multi-level prediction and limits the size of the predicted objects in each level, which greatly alleviates the problem of overlapping bounding boxes and achieves excellent performance.

Interestingly, many one-stage target detectors use two branches to predict position and class separately, such as RetinaNet~\cite{21}, FCOS~\cite{13,22},YoloX~\cite{23}, etc. Sometimes, the branches are not so explicitly divided, and some are needed for auxiliary loss, such as FCOS, which predicts the centroids in the classification branch. Others, such as Mask R-CNN~\cite{25} , add an additional mask branch for task requirements. Even Polar Mask~\cite{26} gives a different meaning to the regression branch in a similar structure to FCOS.

Single-branch nets do not mine relevance between different branches, while multi-branch nets ignore the relevance. But the relevance between different branches is helpful for object detection, so it is necessary to mine and utilize the relevance of different branches.

\section{Approach}

\subsection{Motivation}
The classical single-branch nets can be represented as:
\begin{equation}
\text{class},\text{box} = {f_{fc}}({f_{c2}}({f_{c1}}(\bm{F})),
\end{equation}
where $F$ denotes the input features, $f_{c1}$ and $f_{c2}$ denote the convolutional layers, $f_{fc}$ denotes the fully connected layers. As mentioned above, this leads to the problem of feature coupling.

The classical multi-branch nets can be represented as:
\begin{equation}
\begin{split}
\text{class} = {f_{c3}}({f_{c2}}({f_{c1}}(\bm{F}))),\\
\text{box} = f_{c3}^{\prime}(f_{c2}^{\prime}(f_{c1}^{\prime}(\bm{F}))),
\end{split}
\end{equation}
it uses two branches and does not share parameters, solving the feature coupling problem. However, this approach only emphasizes the differences between branches and ignores the relevance.

How to use the relevance between different semantics is a very challenging task. Considering that the semantics in different branches are not the same and cannot be fused together directly. It is necessary to align the different semantics to explore the common features. Then feature fusion is applied to collect common features of different branches. Similarly, the semantics of the fused features is different from the semantics in the  branch. Therefore, semantic separation is needed to reduce the fused semantics to the original semantics. 

Based on the above analysis, we propose MSIL that integrates semantic alignment, semantic fusion and semantic separation.

\subsection{Network Architecture}
After the definition and description of the problem are given, we propose a simple yet effective solution to the problem called the multi-semantic interactive learning for object detection, and its structure is shown in Figure \ref{fig:net}. The net was divided into three parts: semantic alignment, semantic fusion, and semantic separation.

\textbf{Semantic alignment.} Semantic alignment to extract common features of different branches. In Figure \ref{fig:net}, there are two branches, class, and box branches, which predict class and location, respectively. First the features extracted in the FPN pass through a convolutional layer, which may contain zero or more convolutions inside depending on the net. We call the features after convolution of class branch and box branch $\bm{F}_{cls}$ and $\bm{F}_{reg}$, respectively. Then $\bm{F}_{cls}$ and $\bm{F}_{reg}$ passed through a shared encoder and the results are added with $\bm{F}_{cls}$ and $\bm{F}_{reg}$ to obtain $\bm{F}_{(cls-align)}$ and $\bm{F}_{(reg-align)}$. It is worth noting that the residual structure is formed here, which allows the gradient to be more easily passed back to the lower layer~\cite{27}.

The semantic alignment can be expressed by the formula as follows:
\begin{equation}
\begin{split}
{\bm{F}_{reg-align}} = {f_{conv}}({f_{enc2}}({\bm{F}_{reg}} + {f_{enc}}({\bm{F}_{reg}})),\\
{\bm{F}_{cls-align}} = {f_{conv}}({f_{enc2}}({\bm{F}_{cls}} + {f_{enc}}({\bm{F}_{cls}})),
\end{split}
\end{equation}
where $\bm{F}_{reg}$ and $\bm{F}_{cls}$ denote regression features and classification features, respectively. The aligned feature $\bm{F}_{(reg-align)}$ and $\bm{F}_{(cls-align)}$ were obtained after convolution $f_{env}$,$f_{env2}$ and $f_{conv}$ respectively.

\textbf{Semantic fusion.} Semantic fusion fuses the alignment semantic information in different branches. The independent branches are merged together and passed to semantic separation to extract the relevant information. Semantic fusion concatenates $\bm{F}_{(cls-align)}$ and $\bm{F}_{(reg-align)}$ to obtains $\bm{F}_{fusion}$ by convolution.

The semantic fusion can be expressed by the formula as follows:
\begin{equation}
\begin{split}
{\bm{F}_{fusion}} = {f_{fusion}}(Conc({\bm{F}_{reg-align}},{\bm{F}_{cls-align}})),
\end{split}
\end{equation}
it is concatenated using the function Conc and passed through $\bm{F}_{fusion}$ to obtain the fused features $\bm{F}_{fusion}$,

\begin{figure}[t]
\begin{center}
   \includegraphics[width=0.9\linewidth]{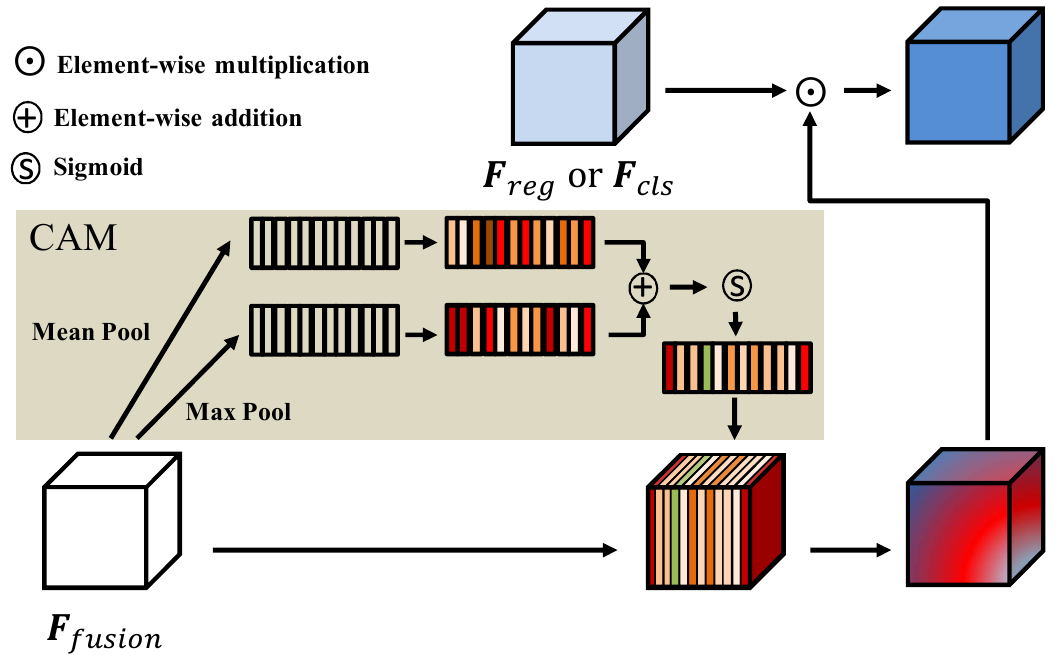}
\end{center}
   \caption{The structure of semantic separation.}
   \label{fig:cam}
\end{figure}
\textbf{Semantic separation.} Semantic separation was applied to reduce the fused semantics to the original semantics, and obtain multi-semantic enhancement features. The net structure of semantic separation as shown in Figure \ref{fig:cam}. Different from the shared encoder in semantic alignment, each branch uses its own independent decoder. In addition, $\bm{F}_{fusion}$ passes through the Channel Attention Module (CAM)~\cite{28} for feature equalization before entering the decoder. The CAM works before the decoder and does not share parameters between branches. Because the attention points are not the same for different tasks. The semantic separation can be expressed by the formula as follows:
\begin{equation}
\begin{split}
\text{class} = {\bm{F}_{cls}}*\sigma {({f_{dec}}(CAM({\bm{F}_{fusion}})*{\bm{F}_{fusion}}) },\\
\text{box} = {\bm{F}_{reg}}*\sigma {(f_{dec}^\prime(CA{M^\prime}({\bm{F}_{fusion}})*{\bm{F}_{fusion}})},
\end{split}
\end{equation}
where the fused features $\bm{F}_{fusion}$ were passed through the CAM to obtain the channel attention. Then the channel attention was multiplied with $\bm{F}_{fusion}$ and passed through to the decoder $f_{dec}$ and the activation function $\sigma$ to obtain the pixel attention. Finally, it is applied to the $\bm{F}_{reg}$ and $\bm{F}_{cls}$.

CAM uses global average pooling and global maximum pooling, respectively, and then combines the information from both to obtain the channel attention, as shown in the following formula:
\begin{equation}
CAM(\bm{F}) = \sigma(MLP(Avg(\bm{F}) + MLP(Max(\bm{F}))),
\label{eq:cam}
\end{equation}
the reason for this design is that the average pooling and maximum pooling will focus on different aspects of the features, and combining them can provide richer information ~\cite{28}.

\subsection{Losses}
There are numerous implementations of object detection, but two tasks are the core issues: predicting the location and the class. MSIL can be integrated into most existing methods. This section shows how to integrate MSIL into other object detection methods with different losses. The classification loss can be expressed as the following formula:
\begin{equation}
L_{cls}(\mathrm{p}_{x, y},\mathrm{t}_{x, y})=\frac{1}{N_{\mathrm{pos}}} \sum_{x, y} L_{\mathrm{ds}}(\mathrm{c}_{x, y}, c_{x, y}^{*}),\\ 
\end{equation}
where $N_{pos}$ denotes the number of predictions, $L_{cls}$ denotes the classification loss, such as Cross-Entropy~\cite{29} loss, Focal Loss~\cite{21}, etc. $c_{x,y}$ denotes the predicted class and $c_{x,y}^*$ denotes the real class. Although the representation of the position differs in the different methods, it is always converted into coordinates, so the regression loss can be expressed as the following formula:
\begin{equation}
L_{reg}(\mathrm{p}_{x, y},\mathrm{t}_{x, y})=\frac{1}{N_{\mathrm{pos}}} \sum_{x, y} \mathbb{Z}_{\{c_{x, y}^{*}>0\}} L_{\mathrm{reg}}(\mathrm{b}_{x, y}, b_{x, y}^{*}),
\end{equation}
where $\mathbb{Z}$ is the indicator function that is 1 when the regulation is true and 0 otherwise. $L_{reg}$ denotes the regression loss, such as intersection over union (IoU)~\cite{30} loss, generalized intersection over union (GIoU)~\cite{31} loss, etc. $b_{x,y}$ denotes the predicted of the position and $b_{x,y}^*$ denotes the real position.The third term denotes the auxiliary loss, which may have one, or more or even none, depending on the method. It can be expressed as the following formula:
\begin{equation}
L_{aug}(\mathrm{p}_{x, y},\mathrm{t}_{x, y})=\sum_{i}^{n} \lambda_{\mathrm{i}} \frac{1}{N_{\mathrm{pos}}} L_{\mathrm{aug}}(\mathrm{u}_{x, y}, u_{x, y}^{*}),
\end{equation}
where $\lambda_i$ denotes the weight of each auxiliary loss, $L_{aug}$ denotes the auxiliary loss, such as center-ness~\cite{22} loss, etc. $u_{x,y}$ and $u_{x,y}^*$ denote the predicted and real values, respectively.

The total loss can be expressed as the following formula:
\begin{equation}
\begin{split}
L(\mathrm{p}_{x, y},\mathrm{t}_{x, y})&=L_{reg}(\mathrm{p}_{x, y},\mathrm{t}_{x, y}),\\
&+{\lambda}L_{cls}(\mathrm{p}_{x, y},\mathrm{t}_{x, y}),\\
&+L_{aug}(\mathrm{p}_{x, y},\mathrm{t}_{x, y}),
\end{split}
\end{equation}
where $\lambda$ denotes the weight, which is used to adjust the percentage of classification loss and regression loss.

\section{Experiments}
MSIL is designed to be integrated into a variety of multi-branch object detection methods. Particularly, we integrate MSIL into FCOS and VFNet respectively to demonstrate the performance of MSIL. In these experiments, all models were trained on a single computer with eight 3080 GPUs on two datasets, MS COCO~\cite{32} and Pascal VOC 2012~\cite{33,34}. MS COCO is obtained by Microsoft on the Amazon Mechanical Turk platform with 80 categories. Pascal VOC 2012 is smaller to MS COCO with 20 categories. 
\subsection{Ablation Study}
We performed a series of ablation experiments to demonstrate the effectiveness of MSIL.

\textbf{Experiment Setting.} MS COCO  train2017 split (115K images) was used for training. Unless specified, ResNet-50 was used as the backbone networks in all ablation experiments. Specifically, our net was trained with stochastic gradient descent (SGD) for 12 epochs with the initial learning rate being 0.01 and a minibatch of 16 images. The learning rate decreases by factor of 0.1 at the 8 and 11 epochs. We initialize our backbone networks with the weights pretrained on ImageNet~\cite{35}.

\begin{table}
\begin{center}
  \begin{tabular}{|c|ccc|}
    \hline
    Method & AP & AP$_{50}$ & AP$_{75}$ \\
    \hline\hline
	w/o semantic alignment & 41.7 & 59.6 & 45.2\\
	w/o semantic separation & 41.9 & 59.8 & 45.3\\
	baseline & \bf{42.1} & \bf{59.9} & \bf{46.0}\\
    \hline
  \end{tabular}
\end{center}
  \caption{Influence of semantic alignment and semantic separation.}
  \label{table:semantic}
\end{table}
\textbf{The influence of semantic alignment and semantic separation.} This experiment explores the influence of semantic alignment and semantic separation. Semantic alignment was used to obtain common features of different branches, semantic separation reduces the fused features to the original semantics. As shown in Table \ref{table:semantic}, the baseline denotes both semantic alignment and semantic separation are adopted in this net. The average precision (AP) without semantic alignment decreased by 0.4\% compared to the baseline, and the AP without semantic separation is 0.2\% lower than the AP of baseline.

It is easy to observe that semantic alignment has a greater influence on the results compared to semantic separation. Because different branches have different semantic information, direct fusion of semantics without alignment can cause confusion. Therefore, it is necessary to extract common features between different semantics through semantic alignment.

\begin{figure}[t]
\begin{center}
   \includegraphics[width=1\linewidth]{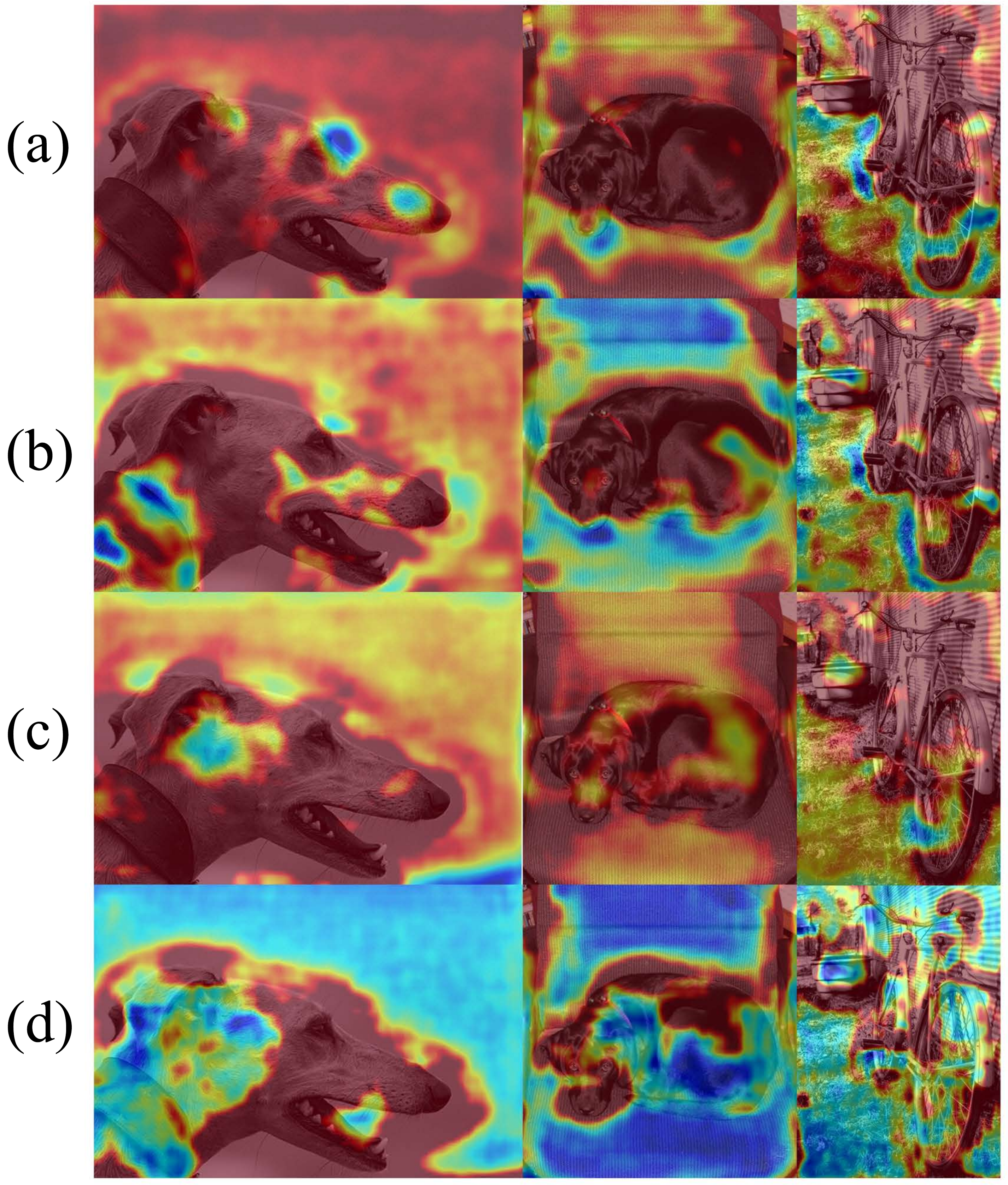}
\end{center}
   \caption{The heat maps of different methods. (a) The heat maps of the classification features of FCOS. (b) The heat maps of the classification features of FCOS+MSIL. (c) The heat maps of the regression features of FCOS. (d) The heat maps of the regression features of FCOS+MSIL.}
   \label{fig:vi}
\end{figure}

 \begin{figure}[t]
\begin{center}
   \includegraphics[width=0.9\linewidth]{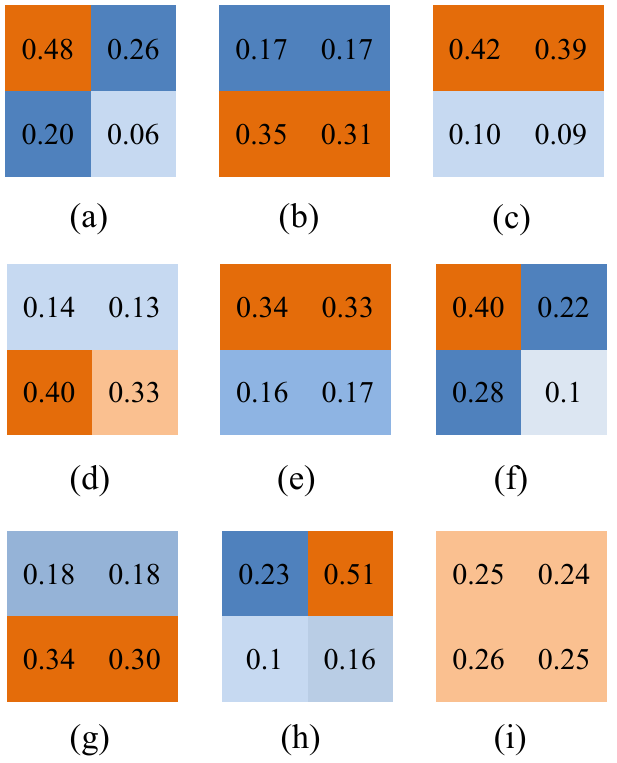}
\end{center}
   \caption{Distributions of object centers, (a) bed, (b) hair drier, (c) skis,  (d) microwave, (e) bicycle, (f) dining table, (g) umbrella, (h) mouse, (i) person.}
   \label{fig:center}
\end{figure}

\begin{figure}[t]
\begin{center}
   \includegraphics[width=0.9\linewidth]{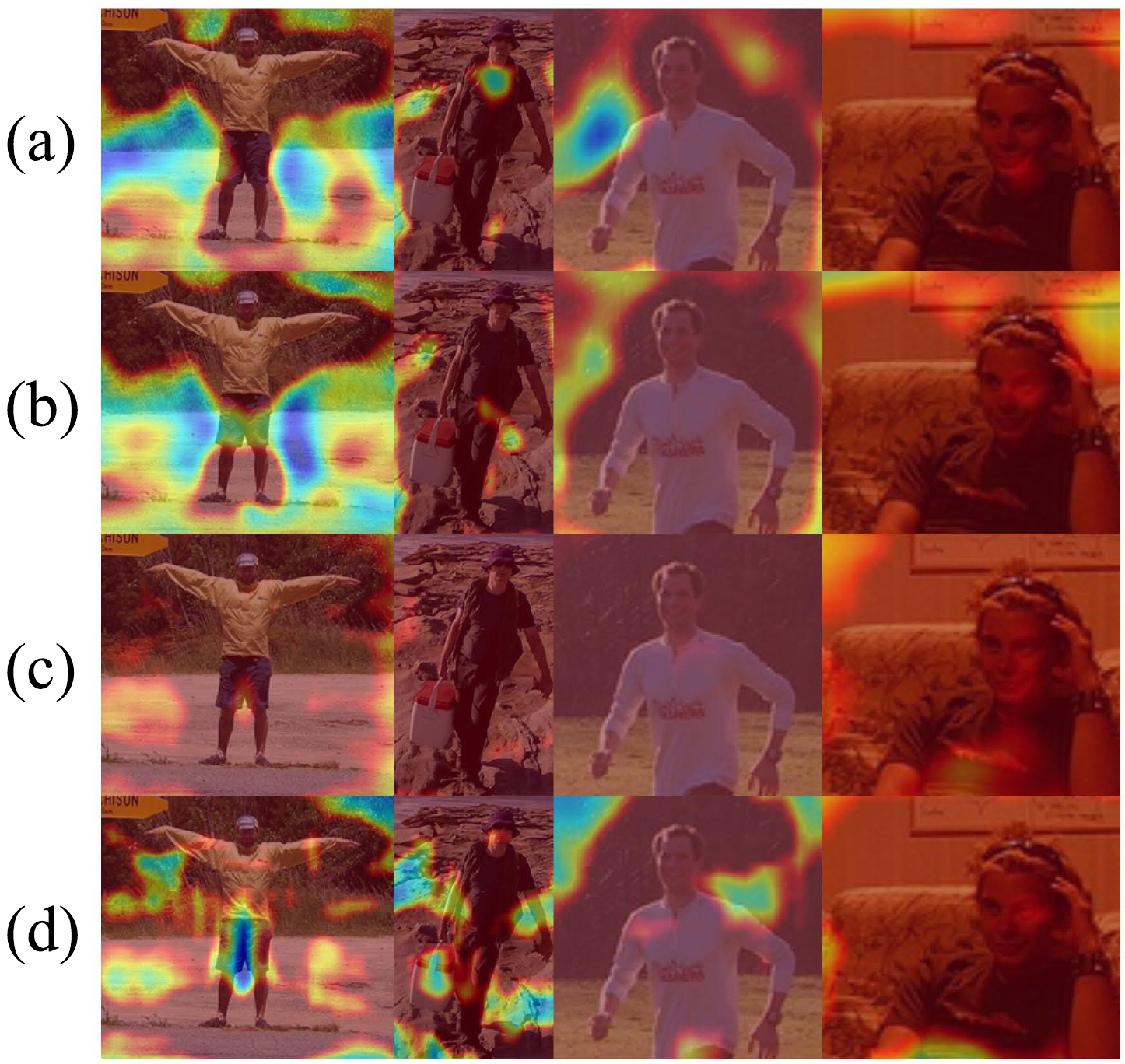}
\end{center}
   \caption{The heat maps of different methods for the person class. (a) The heat maps of the classification features of FCOS. (b) The heat maps of the classification features of FCOS+MSIL. (c) The heat maps of the regression features of FCOS. (d) The heat maps of the regression features of FCOS+MSIL.}
   \label{fig:person}
\end{figure}

\textbf{Feature visualization.} In this experiment, the heat maps is illustrated to show the degree of the image region concerned by the extracted features. The results of FCOS and FCOS+MSIL are shown in Figure \ref{fig:vi}. Blue to red, indicates the attention from low to high.

	In Figure \ref{fig:vi}, (a) and (c) are the heat maps of FCOS multiple branches, (b) and (d) are the heat maps of interactive multiple branches. Comparing (a) and (b) in Figure \ref{fig:vi}, it can be clearly observed that on the classification branch, the red area of FCOS+MSIL is more concentrated in the area where the object is located compared to FCOS. While FCOS more concerned with the entire object, and its red areas tend to cover the full image.

Differently, the red area of FCOS+MSIL is more concentrated in the area where the object is located compared to FCOS. Utilizing the relevant information between branches enables the classification net to better recognize objects.

Comparing (c) and (d) in Figure \ref{fig:vi}, it can be observed that on the regression branches, the red area of FCOS+MSIL is more concentrated to the edge part of the object compared to FCOS, which is more beneficial for regression localization. Based on the observation from the figures of feature visualization, the relevant information is very helpful for the object detection task.

However, not all results are similar. We divided the images into four parts: upper-left, lower-left, upper-right, and lower-right, then counted the distribution of object centers in the COCO training dataset and the results are shown in Figure \ref{fig:center}. It can be observed that there are variations in the distribution of different classes: some classes are more concentrated in the upper left, some are concentrated in the lower part, and some are concentrated in the upper part. However, few classes, such as person, are almost uniformly distributed in the four regions.

Due to the specificity of the person class,we demonstrate the heat maps of the person class in Figure \ref{fig:person}. In this case, the regression branch may not provide additional useful information. It is observed that the region differences that MSIL and FCOS focus on are very limited. MSIL utilizes relevant information to improve performance, which has significant improvements for classes with stronger semantics relevance, and less impact for classes with weaker semantics relevance.

\begin{table}
\begin{center}
  \begin{tabular}{|cc|ccc|}
    \hline
    cls & reg & AP & AP$_{50}$ & AP$_{75}$ \\
    \hline\hline
	\ding{56} & \ding{56} & 41.2 & 59.1 & 44.8\\
	\ding{52} & \ding{56} & 41.9 & 59.9 & 45.6\\
	\ding{56} & \ding{52} & 40.9 & 58.9 & 44.1\\
	\ding{52} & \ding{52} & \bf{42.1} & \bf{59.9} & \bf{46.0}\\
    \hline
  \end{tabular}
\end{center}
  \caption{Feature enhancement of different branches compared with baseline.}
  \label{table:reg_cls}
\end{table}
\begin{table*}
\begin{center}
  \begin{tabular}{|c|c|cccccc|}
    \hline
    Method & Backbone & AP & AP$_{50}$ & AP$_{75}$ & AP$_S$ & AP$_M$ & AP$_L$\\
    \hline\hline
Faster R-CNN w/ FPN~\cite{9} & ResNet-101-FPN & 36.2 & 59.1 & 39.0 & 18.2 & 39.0 & 48.2\\
Faster R-CNN by G-RMI~\cite{36} & Inception-ResNet-v2~\cite{37} & 34.7 & 55.5 & 36.7 & 13.5 & 38.1 & 52.0\\
Faster R-CNN w/ TDM~\cite{38} & Inception-ResNet-v2-TDM & 36.8 & 57.7 & 39.2 & 16.2 & 39.8 & 52.1\\
SSD513~\cite{39} & ResNet-101-SSD & 31.2 & 50.4 & 33.3 & 10.2 & 34.5 & 49.8\\
DSSD513~\cite{40} & ResNet-101-DSSD & 33.2 & 53.3 & 35.2 & 13.0 & 35.4 & 51.1\\
RefineDet~\cite{41} & ResNet-101 & 36.4  & 57.5 & 39.5 & 16.6 & 39.9  & 51.4\\
RetinaNet~\cite{21} & ResNet-101 & 39.1 & 59.1 & 42.3 & 21.8 & 42.7 & 50.2\\
CornerNet~\cite{11} & Hourglass-104 & 40.5 & 56.5 & 43.1 & 19.4 & 42.7 & 53.9\\
FreeAnchor~\cite{42} & ResNet-101 & 43.1  & 62.2  & 46.4  & 24.5  & 46.1 & 54.8\\
SNIP~\cite{43} & ResNet-101 & 43.4  & 65.5 & 48.4 & 27.2 & 46.5  & 54.9\\
SAPD~\cite{44} & ResNet-101 & 43.5 & 63.6 & 46.5 & 24.9 & 46.8 & 54.6\\
ATSS~\cite{45} & ResNet-101 & 43.6 & 62.1 & 47.4 & 26.1 & 47.0 & 53.6\\
MAL~\cite{46} & ResNet-101 & 43.6 & 62.8 & 47.1 & 25.0 & 46.9  & 55.8\\
PAA~\cite{47} & ResNet-101 & 44.8  & 63.3  & 48.7  & 26.5  & 48.8 & 56.3\\
GFL~\cite{48} & ResNet-101 & 45.0 & 63.7  & 48.9  & 27.2  & 48.8  & 54.5\\
OTA~\cite{49} & ResNet-101 & 45.3 & 63.5 & 49.3 & 26.9 & 48.8 & 56.1\\
BorderDet~\cite{50}& ResNet-101 & 45.4 & 64.1 & 48.8  & 26.7 & 48.3  & 56.5\\
GFLV2~\cite{li2021generalized}& ResNet-101 & 46.2 & 64.3 & 50.5  & 27.8 & 49.9  & 57.0\\
DW~\cite{Li_2022_CVPR}& ResNet-101 & 46.2 & 64.8 & 50.0  & 27.1 & 49.4  & 58.5\\
TOOD~\cite{feng2021tood}& ResNet-101 & 46.7 & 64.6 & 50.7  & 28.9 & 49.6  & 57.0\\
FCOS & ResNet-101 & 41.5 & 60.7 & 45.0 & 24.4 & 44.8 & 51.6\\
FCOS+MSIL & ResNet-101 & 42.5 & 61.7 & 46.1 & 25.7 & 45.8 & 53.5\\
VFNet & ResNet-101 & 46.5 & 64.8 & 50.5 & 27.8 & 50.1 & 56.9\\
VFNet+MSIL & ResNet-101 & \textbf{47.1} & \textbf{65.4} & \textbf{51.2} & \textbf{28.6} & \textbf{50.7} & \textbf{58.4}\\
    \hline
  \end{tabular}
\end{center}
  \caption{FCOS+MSIL and VFNet+MSIL vs. other state-of-the-art detectors on coco test-dev.}
  \label{table:sota}
\end{table*}
\textbf{The influence of regression and classification interactivity.} VFNet is applied to investigate the influence of interaction between regression and classification, and the metrics are shown in Table \ref{table:reg_cls}. \ding{52} indicates that the fused features was applied to this branch and \ding{56} indicates not. The AP improved by 0.7\% compared to traditional multi-branch method after applying to the classification branch, indicating that the regression information was helpful for classification. However, applying to regression branch brought counter-intuitive results, with the AP reduced by 0.3\% compared to traditional multi-branch method. 

While feature visualization indicates the regression branch focuses on the edges of the object with classification information,the classification branch may extract the features of the edges instead of the object without location information. Unfortunately, the edge features are not enough to distinguish the object, leading to performance degradation.
On the contrast, the classification branch focuses on the object itself with localization information. Even without the aid of classification information, the regression branch can extract the features of the object region, which is also helpful for localization.

\begin{table}
\begin{center}
  \begin{tabular}{|c|ccc|}
    \hline
    Backbone & AP & AP$_{50}$ & AP$_{75}$ \\
    \hline\hline
	RestNet-50 & 41.2 & 59.1 & 44.8\\
	RestNet-101 & \bf{43.6} & \bf{61.5} & \bf{47.5}\\
    \hline
  \end{tabular}
\end{center}
  \caption{AP with different backbones of VFNet+MSIL.}
  \label{table:backbone}
\end{table}

\textbf{Cooperate with different backbones.} To further demonstrate the effectiveness of our method, we provide experiments with larger backbones. The detailed results are reported in Table \ref{table:backbone}. We evaluate the performances of VFNet+MSIL with the ResNet-50 and ResNet-101 with identical configurations. On the MS COCO val dataset, RestNet-101 improves the AP by 2.4\% compared to RestNet-50.

\subsection{Comparisons with the State of the Art on COCO}
 We integrated MSIL with FCOS and VFNet, respectively, and compared with state-of-the-art detectors on test-dev split of MS-COCO benchmark, as shown in Table \ref{table:sota}. SGD is used for 24 epochs, batch size is 16, the initial learning rate is set to 0.01, weight decay and momentum are set as 1e-4 and 0.9. The learning rate decreases by factor of 0.1 at the 16 and 22 epochs. When integrated with FCOS, our method obtains 1.0\% AP gain. When integrated with VFNet, the AP gain is 0.6\%. It shows that MSIL can utilize the relevance between branches to obtain better performance. Meanwhile, compared to GFLv2 and Borderdet, which are more concerned with regression information, MSIL utilizes more adequate information, resulting in better performance. In addition, MSIL has some advantages compared to TOOD.
\begin{table}
\begin{center}
  \begin{tabular}{|c|c|c|}
    \hline
    Method & Backbone & AP  \\
    \hline\hline
	VFNet & RestNet-50 & 65.3\\
	VFNet+MSIL & RestNet-50 & 66.4\\
	FCOS & RestNet-50 & 71.0\\
	FCOS+MSIL & RestNet-50 & \bf{71.8}\\
    \hline
  \end{tabular}
\end{center}
  \caption{FCOS+MSIL and VFNet+MSIL vs FCOS and VFNet on the VOC dataset.}
  \label{table:voc}
\end{table}

\subsection{FGPL on Pascal VOC}
We also compared on VOC dataset, using VOC 2012 train as the training set and evaluated on VOC 2012 val set, as shown in Table \ref{table:voc}. The training was done for 24 epochs and the other settings were the same as the configurations mentioned above.

FCOS+MSIL is 0.8\% higher compared to FCOS at the same backbone and VFNet+MSIL is 1.1\% higher compared to VFNet. It is further demonstrated that MSIL can utilize the relevance between branches to obtain better performance.

\section{Conclusion}
This paper proposes multi-semantic interactive learning for object detection (MSIL). MSIL extracts relevance between different branches by semantic alignment, semantic fusion and semantic separation. The experimental results show that the multi-semantic enhanced features of the classification focus more on the object region and  the multi-semantic enhanced features of the regression focus more on the boundary of the object. More importantly, MSIL can be integrated into other multi-branch net. Quantitative experimental results show that MSIL achieves the better numerical performance on both MS COCO and Pascal VOC datasets.

{\small
\bibliographystyle{ieee_fullname}
\bibliography{egbib}
}

\end{document}